\pdfoutput=1

\documentclass[11pt]{article}

\usepackage{acl}

\usepackage{times}
\usepackage{latexsym}

\usepackage[T1]{fontenc}

\usepackage[utf8]{inputenc}

\usepackage{microtype}

\usepackage{xcolor}
\usepackage{algorithm}
\usepackage{algpseudocode}
\usepackage{pifont}
\usepackage{amsmath}
\usepackage{graphicx}
\usepackage{amsmath,amsfonts,amssymb}
\usepackage[inline]{enumitem}
\usepackage{arydshln}

\title{LexSubCon: Integrating Knowledge from Lexical Resources into  Contextual Embeddings for Lexical Substitution}

\author{%
George Michalopoulos, Ian McKillop, Alexander Wong,  Helen Chen\\
    University of Waterloo, \\  Waterloo, Canada \\
\{gmichalo, ian,  alexander.wong, helen.chen\}@uwaterloo.ca
}

\begin{document}
\maketitle
\begin{abstract}
Lexical substitution is the task of generating meaningful substitutes for a word in a given textual context. Contextual word embedding models have achieved state-of-the-art results in the lexical substitution task by relying on contextual information extracted from the replaced word within the sentence.  
However, such models do not take into account structured knowledge that exists in external lexical databases.

We introduce LexSubCon, an end-to-end lexical substitution framework based on contextual embedding models that can identify highly-accurate substitute candidates. This is achieved by combining contextual information with knowledge from structured lexical resources. Our approach involves: \begin{enumerate*}[label=(\roman*)] \item introducing  a novel mix-up embedding strategy to the target word's embedding through linearly  interpolating the pair of the target input embedding and the average embedding of its probable synonyms; \item  considering the similarity of the sentence-definition embeddings of the target word and its proposed candidates; and, \item  calculating the effect of each substitution on the semantics of the sentence through a fine-tuned sentence similarity model. \end{enumerate*}  Our experiments show that LexSubCon outperforms previous state-of-the-art methods by at least 2\% over all the official lexical substitution metrics on LS07  and CoInCo benchmark datasets that are widely used for lexical substitution tasks.

\end{abstract}

\section{Introduction}

Lexical Substitution \cite{mccarthy-navigli-2007-semeval} is the task of generating appropriate words which can replace a target word in a given sentence without changing the sentence's meaning. The increased research interest in Lexical Substitution is due to its utility in various Natural Language Processing (NLP) fields including data augmentation, paraphrase generation and  semantic text similarity.

 Contextual word  embedding  models (such  as  ELMo   \cite{peters-etal-2018-deep} and BERT \cite{devlin-etal-2019-bert}) have achieved state-of-art results in many NLP tasks. These models are usually pre-trained on massive corpora and the resulting context-sensitive embeddings are used in different downstream tasks \cite{howard-ruder-2018-universal}.  \citet{zhou-etal-2019-bert} have achieved state-of-the-art results on the lexical substitution task by improving the BERT's standard procedure of the masked language modeling task. However, the current state-of-the-art contextual models have yet to incorporate   structured knowledge that exists in external lexical database into their  prediction process. These lexical resources  could boost the model's performance by providing additional information  such as the definitions of the target   and  candidate words (in order to ensure that the candidate word is semantically similar to the target word and not only appropriate for the sentence's context) or by  enriching the proposed candidate word list so it will not only be based on the vocabulary of the contextual model.
 
In this paper,  we present and publicly release\footnote{https://github.com/gmichalo/LexSubCon} 
a novel framework for the lexical substitution task. Specifically,
\begin{enumerate*}[label=(\roman*)]
\item we are the first, to the best of our knowledge, to propose a novel  mix-up embedding strategy that outperforms the previous state-of-the-art strategy of word embedding dropout for the input embedding of the target word in a contextual model \cite{zhou-etal-2019-bert} for the task of predicting accurate candidate words;
\item we  propose the combined usage of  features from contextual embedding models and external lexical knowledge bases in order to determine the most appropriate substitution words without modifying the meaning of the original sentence, such as introducing a new gloss (definition) similarity metric which calculates  the similarity of the sentence-definition embeddings of the target word and its proposed candidates;
\item we generate a highly accurate fine-tuned sentence similarity model by taking advantage of popular data augmentation techniques (such as back translation), for calculating the effect of each candidate word in the semantics of the original sentence; and,
\item finally, we show that LexSubCon achieves state-of-the-art results on two popular benchmark lexical substitution datasets \cite{mccarthy-navigli-2007-semeval, kremer-etal-2014-substitutes}.
\end{enumerate*}


\section{Related Work}
\label{related}

The lexical substitution task consists of two sub-tasks: \begin{enumerate*}[label=(\roman*)] \item  generating a set of meaning preserving substitute candidates for the target word  and \item appropriately ranking  the words of the set by their ability to preserve the meaning of the initial sentence \cite{giuliano-etal-2007-fbk-irst, martinez-etal-2007-melb}. \end{enumerate*} However, lexical substitution models can also be tested in a ``simpler'' problem where the set of substitute candidates is composed of  human-suggested  words and the task is to accurately rank the substitute words that are provided \cite{erk-pado-2010-exemplar}.

The authors in \cite{melamud-etal-2015-simple}  proposed the use of a word2vec model which  utilizes word and context embeddings to represent  the target word in a given context. Their model  ranked the candidate substitutions by measuring their embedding similarity. In \cite{melamud-etal-2016-context2vec} the context2vec model was introduced where the context representation of the word was calculated by combining the output of two bidirectional LSTM models using a feed-forward neural network. 


 \citet{peters-etal-2018-deep} introduced contextualized word embeddings in a bidirectional language model (ELMo). This allowed the model to change the embedding of a word based on its imputed meaning which is derived from the surrounding context. 
Subsequently, \citet{devlin-etal-2019-bert} proposed the  Bidirectional  Encoder  Representations  from  Transformers  (BERT)  which uses bidirectional transformers \cite{Vaswani2017AttentionIA} to create context-dependent representations. 
The authors in \cite{gari-soler-etal-2019-comparison} used ELMo in the lexical substitution task by calculating the cosine similarity between the  ELMo embedding of the target word and all the candidate substitutes. In addition, \citet{zhou-etal-2019-bert} achieved state-of-the-art  results on the lexical substitution task by applying  a dropout embedding policy to the target word embedding and by taking into account the similarity between the  initial contextualized representation of the context words and their representation after replacing the target word by one of the possible candidate words. An  analysis of state-of-the-art 
contextual model   on the lexical substitution task was presented in  \cite{compa_study}.

Finally, external knowledge from  knowledge bases has  been used to enhance the  performance of deep learning models. Sense-BERT \citep{Levine2020SenseBERTDS}  was pre-trained to predict the semantic class of each word by incorporating lexical semantics (from the lexical database WordNet \citep{10.1145/219717.219748}) into the model's pre-training objective. Furthermore, 
\citet{faruqui-etal-2015-retrofitting} and \citet{Bahdanau2017LearningTC} used external knowledge (namely WordNet) in order to enhance word embeddings  and to create more accurate representations of rare words.

\section{LexSubCon Framework}


\label{methods}

In the lexical substitution task, a model aims to firstly  generate a set of candidate substitutions for each target word and secondly to create an appropriate  ranking of the elements of the candidate set. In addition,  
there are two main conditions for a lexical substitute model to satisfy: \begin{enumerate*}[label=(\roman*)] \item  to be semantically similar to the target word and \item  and to be compatible with the given context (sentence) \cite{melamud-etal-2015-simple}.
\end{enumerate*}
We present the LexSubCon framework   which achieves state of the art results on the lexical substitution task  by combining contextual information with knowledge from structured external lexical resources. 

\begin{figure}[h]
        \centering
        \includegraphics[width=\linewidth]{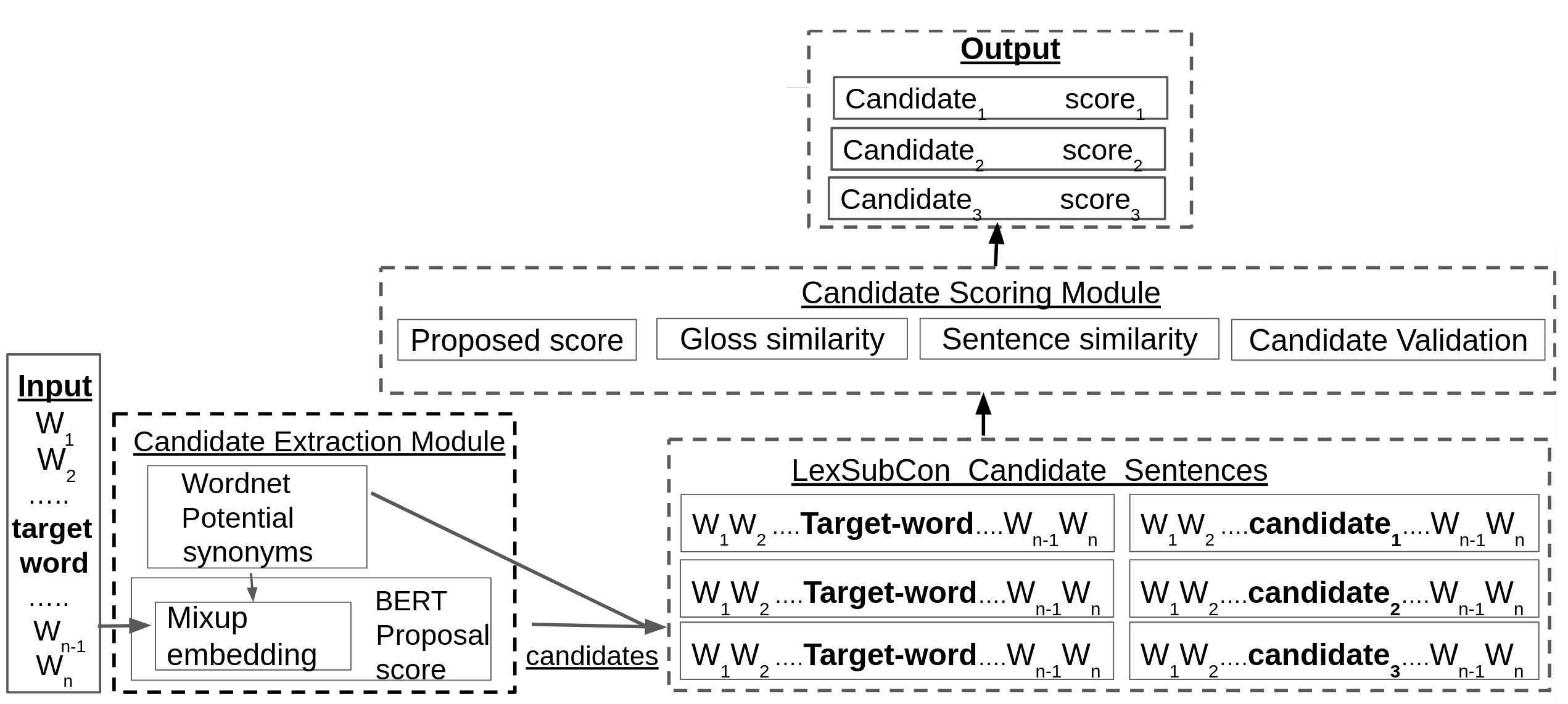}
    \caption{LexSubCon framework }
    \label{fig:1}
\end{figure}
The architecture of LexSubCon is depicted in Figure \ref{fig:1}.
The key characteristic of  LexSubCon is its capability of unifying different substitution  criteria such as   contextualized representation, definition and sentence similarity  in a  single framework  in order to accurately identify suitable candidates for the target words in a specific context (sentence).

\subsection{Proposed Score: Mix-Up Embedding Strategy}
\label{sec:bert-mix}

The standard BERT architecture \cite{devlin-etal-2019-bert}  can be used in the lexical substitution task by masking the target word and letting the model to propose appropriate substitute candidates that preserve the initial meaning of the sentence.   \citet{zhou-etal-2019-bert} argued that applying embedding dropout to partially mask the target word is a better alternative than masking the whole word. This is because  the model may generate candidates that are semantically different but appropriate for the context of the initial 
sentence. Their experiments showed that this policy is indeed more beneficial than completely masking, or not masking, the target word.

However, in this paper we  demonstrate that a mix-up embedding strategy can yield even better results. The main disadvantage of  dropout embedding is that it sets random positions in the embedding vector of the target words to zero. 
We propose that by using external knowledge, we can obtain probable synonyms of the target word and use that knowledge in a mix-up scenario \cite{Zhang2018mixupBE}  through linearly interpolating the pair of the target input embedding and the average embedding of its synonyms. This allows the model to generate a new synthetic input embedding by re-positioning the target embedding around the neighborhood of the embedding of its synonyms. 
In order to obtain appropriate synonyms we use  WordNet \cite{10.1145/219717.219748}  which is an extensive lexical database where words are grouped into sets of synonyms (synsets). In our experiments, the best performance was achieved when  the list of synonyms was extracted from the complete set of synsets for each word as it minimizes the chance of having a synonym set that only includes the target word itself.

Finally, we use a mix-up strategy to calculate a new input embedding for the target word $X'_{target}$ as shown in equation \ref{eq:e1}:

\begin{equation}
X^{'}_{target} = \lambda X_{target} + (1- \lambda) \overline{X}_{synonyms}
\label{eq:e1}
\end{equation}
Where $X_{target}$ is the initial input embedding of the target word, $\overline{X}_{synonyms}$ is the average embedding of all the synonyms.  
It should be noted that Wordnet does not contain information about some words, such as pronouns, conjunctions, or nouns that are not commonly used in the English vocabulary. To address this limitation, whenever a target word cannot be found in the WordNet database,  we replace the mix-up strategy by injecting  Gaussian noise to the input embedding of the target word. This produces a similar effect as the mix-up strategy since the target embedding is re-positioned around itself in the embedding space  (equation \ref{eq:2}):

\begin{equation}
X^{'}_{target} = X_{target} + 
e
\label{eq:2}
\end{equation}
where $e$ is a Gaussian noise vector with components $e_{i} \sim \mathcal{N}(\mu_i,\,\sigma^{2}_i)$.

We use the 
BERT architecture to calculate the proposal score for each candidate. The input embedding vectors pass through multiple attention-based transformer layers  where each layer produces a contextualized embedding of each token. For each  target word $x_t$, the model  outputs a score vector  $y_t   \in \mathbb{R}^{D} $, where $D$ is the length of the model's vocabulary. We  calculate the proposal score  $s_p$ for each candidate word $x_c$, using the score vector  $y_t$ of the BERT's language modeling process,   as the  probability for the BERT model to propose the word $x_c$ over all the candidates words $x'_c$  when the target word's sentence is provided as input to it:
\begin{equation}
s_p (x_c) = \frac{exp(y_t [x_c])}{\sum_{x'_c} exp(y_t [x'_c])}
\label{eq:1}
\end{equation}


\subsection{Gloss-Sentence Similarity Score}
\label{sec:gloss}

In the previous section, we analyzed our model which ranks candidate substitute words by calculating their individual proposal scores. However, \citet{zhou-etal-2019-bert} and \citet{compa_study} showed that the proposal score does not provide sufficient information about whether the substitute words will modify the sentence's meaning.
Thus, in this section, we present a new metric which ranks the candidate words by considering the gloss (a dictionary-style definition) of each word. By extracting the appropriate information from the Wordnet database, a list of potential glosses  is created for each target or candidate word. In addition, we can determine the most appropriate gloss based on  the  word and its specific context (sentence) by taking advantage of recent  fine-tuned  contextual models that have achieved state-of-the-art results  in the Word Sense Disambiguation (WSD) task \cite{huang-etal-2019-glossbert}. As  the glosses are sentences (sequence of words) they can  be represented in a semantic space through a sentence embedding generating model. A ranking of each candidate word is produced by calculating the cosine similarity between the gloss sentence embedding of the target word and the gloss sentence embedding of each candidate word. 

There are many methods for generating  sentence embeddings, such as calculating the  weighted average of its word embeddings \cite{Arora2017ASB}. We select the   sentence embeddings of the stsb-roberta-large model  in \cite{sentencebert}  which has been shown to  outperform   other state-of-the-art sentence embeddings methods.

Given a sentence $s$, a target word $x_t$ and a candidate word $x_c$, our model first identifies the most appropriate gloss $g_t$ for the target word given its context. After replacing the target word with the candidate  $x_c$ to create a new sentence $s'$,  the most appropriate gloss  $g_c$ for the candidate word is also determined. A gloss-similarity score $s_g$ for each candidate is then calculated as the cosine similarity between the two glosses-sentences embeddings.

\begin{equation}
s_g(x_c) =  cos(g_t, g_c)
\label{eq:4}
\end{equation}


\subsection{Sentence Similarity Score}
\label{sec:back}

We also chose to calculate the effect of each substitution in the semantics of the original sentence by calculating the semantic textual similarity  between the original sentence $s$ and the updated sentence $s'$ (a sentence where we have replaced the target word with one of its substitutions).

In order to accurately calculate a similarity score  between $s$ and $s'$, we fine-tune a semantic textual similarity model based on the stsb-roberta-large model \cite{sentencebert} by using the training portion of the dataset in order to create pairs of sentences between the original sentence and an updated sentence where we have substitute the target word with one of its proposed candidates. Using  the methods that we described in  section \ref{sec:gloss}, we can identify the most appropriate synset (from WordNet) for each target word and create a new pair of sentences between  the original sentence and an updated sentence where we have updated the target word with the synonyms of the previous mentioned synset. However, due to the limited size of the training dataset, our model is still  not provided with enough training data in order to be fully fine-tuned.

This is the reason why we employ a data augmentation technique in order to produce the examples needed for this task. Specifically, we create a back-translation mechanism in order to generate artificial  training data. Back-translation or round-trip translation is the process of translating text into another language (forward translation) and then translating back again into the original language (back translation) \cite{milam}. Back-translation has been used in different tasks in order to increase the size of training data \cite{sennrich-etal-2016-improving,aroyehun-gelbukh-2018-aggression}. In our case, we  provide to the back-translation module the initial sentence $s$ and it produces a sightly different `updated' sentence $s_u'$. For the $s_u'$ sentences that still contain the target word  we can create  pair of sentences between the $s_u'$  and an alternative version of the $s_u'$ sentence ($s_u''$) where the target word is substituted with one of the candidate words or synonyms that we mentioned in the above paragraph. The main disadvantage of this techniques is that it may return the same initial sentence without any changes. In this case, we add a second translation level where the initial sentence is translated into two different languages before being translated back. 



  \subsection{Candidate Validation Score}
  \label{cad_val}
In our experiments we have also included the substitute candidate validation metric from \cite{zhou-etal-2019-bert} as it  has been shown to have a positive affect on the performance of a lexical substitution model. The substitute candidate validation metric  is represented as  the weighted sum of the cosine similarities between the contextual representation of each token in the initial and in the updated sentence where the weight of the cosine similarity of the token $i$ is calculated as the average self-attention score of all heads in all layers from the token of the target word to token $i$. As  mentioned in \cite{zhou-etal-2019-bert}, this metric  evaluates the influence of the substitution on the semantic of the sentence. 

Finally,   LexSubCon uses a linear combination of the above mentioned features to calculate the final score for each candidate word.

\subsection{Candidate Extraction}
\label{candext}
The candidates for each target word are extracted using the external 
lexical resource of WordNet and the BERT-based lexical substitution approach where the model provides probabilities for each candidate based on the context (sentence). 
We create a list of candidates based on the synonyms, the hypernyms, and hyponyms of each target word that could be identified in WordNet. In addition, we  include in the list the candidate words with the higher probability that could be identified using the mix-up strategy that we described in section \ref{sec:bert-mix}. We chose to  include candidates from WordNet because we do not want our model to only include candidates words from the BERT vocabulary and we also include candidates words from a BERT-based model because target words may not be included in WordNet. 
\section{Experiments}
\label{results}

 \begin{table*}[h!]
 \centering
  \begin{tabular}{l  c c c c c}
  \hline

\textbf{Method}   & best  &     best-m &   oot &    oot-m &  $P@1$   \\ \hline
&&\multicolumn{2}{c}{LS07 dataset}&
\\ \hline
LexSubCon    & \textbf{21.1} $\pm$0.03 &  \textbf{35.5}$\pm$0.07 & \textbf{51.3} $\pm$0.05 &   \textbf{68.6}$\pm$0.05 & \textbf{51.7}$\pm$0.03\\
 Bert$_{s_p, s_u}$*  & 12.8 $\pm$0.02 &     22.1$\pm$0.03 &   43.9$\pm$0.01 &    59.7$\pm$0.02 & 31.7$\pm$ 0.02  \\ 
Transfer learning  &17.2  &  - &   48.8 &    -&    - \\
Substitute vector   & 12.7   &  21.7 &   36.4 &    52.0&    -\\
 Addcos  & 8.1  &  13.4 &   27.4 &    39.1 & -\\
Supervised learning &15.9  & - &   48.8 &    - & 40.8 \\
KU  &12.9  &     20.7 &   46.2 &    61.3 &    -\\
UNT  &12.8  &     20.7 &   49.2 &    66.3 &    -\\
 \hline
&&\multicolumn{2}{c}{CoInCo dataset} &
\\ \hline
LexSubCon    & \textbf{14.0}  $\pm$ 0.02  & \textbf{29.7}  $\pm$ 0.03 &  \textbf{38.0} $\pm$ 0.03 &\textbf{59.2} $\pm$ 0.04 & \textbf{50.5} $\pm$ 0.02\\ 
Bert$_{s_p, s_u}$ *  & 11.8 $\pm$ 0.02 &    24.2 $\pm$ 0.02 & 36.0  $\pm$ 0.02 & 56.8  $\pm$ 0.02 & 43.5  $\pm$ 0.02\\ 

Substitute vector   & 8.1   & 17.4  & 26.7  & 46.2&    -   \\
 Addcos   & 5.6 & 11.9 & 20.0 &33.8 & -  \\\hline
  \end{tabular}
  \caption{Results of mean $\pm$ standard deviation of five runs from our implementation  of LexSubCon  and  Bert$_{s_p, s_u}$\cite{zhou-etal-2019-bert}. We also provide the performance of previous state-of-the-art models. Transfer learning \cite{hintz-biemann-2016-language}, Substitute vector \cite{melamud-etal-2015-modeling},  Addcos  \cite{melamud-etal-2015-simple},  Supervised learning \cite{szarvas-etal-2013-learning}, 
  KU \cite{Dearticle}, UNT \cite{hassan-etal-2007-unt}.  Best values are \textbf{bolded}. }
  \label{tab:res}
\end{table*}

\subsection{Dataset}

We evaluate LexSubCon on the English datasets 
SemEval 2007 (LS07)\footnote{license: https://tinyurl.com/semeval-license} \cite{mccarthy-navigli-2007-semeval} and Concepts-In-Context (CoInCo)\footnote{license: CC-BY-3.0-US }  \cite{ kremer-etal-2014-substitutes}   which are the most widely used datasets for the evaluation of lexical substitution models. 
\begin{enumerate*}[label=(\roman*)]
    \item The LS07 dataset is split into 300 train   and 1710 test sentences where for each of the 201 target words there are 10 sentences (extracted from http://corpus.leeds.ac.uk/internet.html). The gold standard was based on manual annotation where annotators  provided up to 3 possible 
    substitutes.
    \item    The CoInCo dataset  consists of over 15K target word instances (based on texts provided in the Open American National Corpus)  where 35\% are training   and 65\% are testing data. Each annotator provided at least 6 substitutes for each target word. 
Our experiments with all datasets 
are consistent with their intended use, as they were created for research purposes. We  manually investigate the existence of information that names individuals or offensive content, however, we did not find any indication of either of them.
\end{enumerate*}

In order to have a fair comparison with the previous state-of-the-art models, for both datasets we used their processed versions as used in \cite{melamud-etal-2015-simple, melamud-etal-2016-context2vec}.

\subsection{Experimental Setup}
\label{sec:setup}
LexSubCon was evaluated in the following   variations of the lexical substitution tasks:

 \textbf{All-ranking task: } In this task  no substitution candidates are provided. We use the official metrics that the organizers provided in the original lexical substitution task of SemEval-2007 \footnote{www.dianamccarthy.co.uk/files/task10data.tar.gz }. These were
\textit{best} and \textit{best-mode} which validate the quality of the model's best prediction and both \textit{oot} (out-of-ten) and \textit{oot-mode} to evaluate the coverage of the gold substitute candidate list by the 10-top predictions. We also use  $Precision@1$ 
to have a complete comparison with the model in \cite{zhou-etal-2019-bert}.

    \textbf{Candidate ranking task:} In this task the  list of candidates are provided and the goal of the model is to rank  all the candidate words.  For the candidate ranking task we follow the policy of previous works and  construct the candidate list by merging all the substitutions of the target lemma and POS tag over the whole dataset. 
    For measuring the performance of the model  we use the GAP score  \cite{kish}\footnote{https://tinyurl.com/gap-measure}  which is a variant of the MAP (Mean Average Precision). Following \cite{melamud-etal-2015-simple}, 
     we discard all multi-words from the gold substitutes list and remove the instances that were left with no gold substitutes.

We use the uncased BERT  large model \cite{devlin-etal-2019-bert} for the calculation of the proposal score  and candidate validation score. For the identification of the most appropriate glosses 
we employ the pre-trained  model in \cite{huang-etal-2019-glossbert} which has achieved the state-of-the-art results in the Word Sense Disambiguation (WSD) task. Finally,  the sentence-similarity metric is computed by fined-tuning 
the stsb-roberta-large model in \cite{sentencebert} and by employing the  OPUS-MT models \cite{TiedemannThottingal:EAMT2020} (namely opus-mt-en-romance, opus-mt-fr-es and opus-mt-romance-en) 
for the creation of the back-translated sentences. 

We use the LS07 trial set for training the sentence similarity metric model (for 4 epochs)  and for fine-tuning the parameters of our framework based on the \textit{best} score. Empirically,   the $\lambda$ parameter of the mix-up strategy was set to $0.25$ and   the weights to $0.05, 0.05, 1, 0.5$ for the proposal score, gloss-sentence similarity score, sentence similarity score and candidate validation score respectively (with the search space for all the parameters being $[0,1]$\footnote{As we only had four weight parameters, the identification of the best combination  was finished in less than half an hour.}). 
Finally, for the Gaussian noise we choose a mean value of 0 and standard deviation 0.01. We propose 30 candidates for each target word. 
In order to achieve more robust results,  we run LexSubCon on five different (random) seeds  and we provide the average  scores and standard deviation.
All the contextual  models are implemented using the transformers library \citep{Wolf2019HuggingFacesTS} on PyTorch 1.7.1. All experiments are executed on a Tesla K80  GPU with 64 GB of system RAM on Ubuntu 18.04.5 LTS. It should be noted that
 LexSubCon contains 1136209468 parameters.

\subsection{Lexical Substitution Model Comparison}

 To enable direct comparison and to isolate gains due to improvements solely on the post-processing strategy that each model uses (which has the potential to change its performance  \cite{compa_study}), we opt to reproduce and use the same strategy  for the tokenization of the target words from Bert$_{s_p, s_u}$ \cite{zhou-etal-2019-bert}. 
 We focus our comparison on Bert$_{s_p, s_u}$  as it has achieved impressive state of the art results on both  benchmark datasets\footnote{Note that the method proposed by \cite{zhou-etal-2019-bert} was implemented to the best of our abilities to be as faithful to the original work as possible 
 using elements of code that the method's authors kindly provided upon request.  However, the  authors could not make the complete original code available to us.}. 
 
  The results of LexSubCon and the previous state-of-the art results in both LS07 and CoInCo benchmark datasets are presented in  Table \ref{tab:res}.  LexSubCon outperformed the previous methods across all  metrics in the LS07 and the CoInCo datasets 
    due to the fact that all features have a positive contribution on its performance (see ablation details in section  \ref{sec:ablation}) as they encourage LexSubCon to take into consideration different substitution criteria. 
The standard deviation of the results of LexSubCon is not zero due to the fine-tuning process of the sentence similarity model. However, the results indicate that there are  no large fluctuations. 
 LexSubCon  and our implementation of Bert$_{s_p, s_u}$ had a running time of  74k and   30k  for LS07 and 580K and  266K seconds for the  CoInCo dataset respectively.

\subsection{Mix-Up Strategy Evaluation}
\label{result-mixup}

 \begin{table}[h!]
 \centering
  \begin{tabular}{l  c c c c c}
  \hline

\textbf{Policy}   & best  &     best-m &   oot &    oot-m  &$P@1$  \\ \hline
&\multicolumn{2}{c}{LS07} &
\\ \hline
Mix. & \textbf{16.3} & \textbf{27.6} & \textbf{45.6} & \textbf{62.4}& \textbf{40.8}\\
Gaus. &15.4 &   25.1 & 44.3 &  61.4& 38.9 \\
Drop. &  15.5 & 25.6 & 44.3 & 61.2 & 38.8\\
Mask & 10.4 & 16.4 & 35.5 & 48.6 &27.0 \\
Keep & 15.5 & 25.4 & 44.4 &  61.4 & 39.2\\
\hline
&\multicolumn{2}{c}{CoInCo} &
\\ \hline
Mix. & \textbf{11.3} & \textbf{23.8} & \textbf{33.6} &\textbf{54.4} & \textbf{41.3} \\
Gaus. & 10.8 & 22.6 & 33.0 & 54.4 &39.7 \\ 
Drop.  & 10.8 &  22.5& 32.9 & 54.2& 39.5\\
Mask  & 8.6 & 17.5 & 28.9 &  46.6 & 31.7 \\
Keep & 10.8 & 22.6 &  33.0 & 54.3 &  39.7 \\\hline
  \end{tabular}
  \caption{Comparison of different strategies for modifying the input embedding of the proposal model.  \textit{Mix.} is the mix-up strategy that we proposed,  \textit{Gaus.} is the Gaussian noise strategy, \textit{Drop.} is the dropout embedding strategy in \cite{zhou-etal-2019-bert},    \textit{Mask} is the strategy of masking the target word and  \textit{Keep} is the strategy of unmasking the target word in the input of the proposal model. Best values are \textbf{bolded}. }
  \label{tab:res_mixup}
\end{table}

In order to evaluate the mix-up strategy for the input embedding of the proposal model,
we study the effect of different input embedding policies. The results of this study are listed in Table \ref{tab:res_mixup}. It can be observed that  even the simpler strategy of injecting Gaussian noise to the input embedding outperformed the standard policy of masking the input word. 
These results indicate that a contextual model  needs information from the embedding of the target word in order to predict accurate candidates but it may over-rely on this information when it is provided with an intact input embedding. 
Furthermore, the mix-up strategy outperformed all the other policies  and specifically the  dropout embedding strategy  \cite{zhou-etal-2019-bert} as  the mix-up strategy re-positions the target embedding around the neighborhood of the embedding of its synonyms and it does not 
erase a part of the embedding that the model can learn from.

\subsection{Ablation Study}
\label{sec:ablation}
 \begin{table}[h!]
 \centering
  \begin{tabular}{l  c c c c c}
  \hline

\textbf{Method}   & best  &     best-m &   oot &    oot-m &$P@1$   \\ \hline
&\multicolumn{2}{c}{LS07} &
\\ \hline
\textbf{LexS} & \textbf{21.1}  &  \textbf{35.5}  & \textbf{51.3}   &   \textbf{68.6}  &\textbf{51.7} \\
\textit{-w Pr.  } &20.1 & 32.6 & 50.8 & 68.1&50.6 \\ 
\textit{-w Gl.  } &19.9 & 33.7 &  50.4 & 67.6&48.6 \\
\textit{-w Sen.} &20.7& 34.9& 50.9& 68.2&50.6 \\
\textit{-w Val. }&18.8& 31.7&47.8& 64.9& 46.6\\
\hdashline

 \textit{Pr.}& 16.3&27.6&45.6&62.4&40.8\\
 \textit{Gl.}  & 12.4& 19.5 & 40.5 & 55.0&32.7 \\
 \textit{Sen.}& 16.7&28.3&45.3&62.0&40.7\\
 \textit{Val.}  & 18.6 & 30.8 &48.9 & 66.2&46.3 \\  \hline

&\multicolumn{2}{c}{CoInCo} &
\\ \hline
\textbf{LexS} & \textbf{14.0}     & \textbf{29.7}   &  \textbf{38.0}   &\textbf{59.2} &  \textbf{50.5}  \\
\textit{-w Pr.} & 12.9& 26.5& 37.6& 58.5& 47.8\\
\textit{-w Gl.} & 13.4& 28.5 & 37.2& 58.2&48.8\\
\textit{-w Sen. } & 13.6 & 29.9&37.2& 58.3&49.2\\
\textit{-w Val.}& 12.7& 27.0& 35.9& 57.4& 46.6 \\
\hdashline
 \textit{Pr.}  & 11.3 & 23.8& 33.6 & 54.4 & 41.3  \\
 \textit{Gl.}  & 8.4 & 16.7 & 29.6 & 47.2 & 33.6\\
 \textit{Sen.}  & 10.9 & 22.5 & 34.0 & 54.9&40.5\\
 \textit{Val.} & 11.7 & 23.7 & 35.3 & 55.2& 44.2\\

\hline

  \end{tabular}
  \caption{Ablation study of LexSubCon: \textit{Pr.} is the Proposal score using the mix-up embedding strategy. \textit{Gl.}  is the Gloss  similarity score.    \textit{Sen.} is the Sentence Similarity score and  \textit{Val.} is the  Validation score. \textit{-w/o} indicates a LexSubCon framework \textbf{without} the specific feature.}
  \label{tab:abl}
\end{table}

In order to evaluate the effect of each feature on the performance of LexSubCon, we conducted an ablation study. The results are presented in Table \ref{tab:abl}. As Table \ref{tab:abl} shows, LexSubCon achieved its best performance when it has access to information  from all the features  described in section \ref{methods}. 
By testing the performance of the individual features, we observe that the gloss sentence similarity feature achieves the worst performance out of all the features. This is likely  because many candidate words cannot be identified in Wordnet and thus we assign a zero value to their gloss sentence score.  Another factor is that the models that were used for the selection of the most appropriate gloss for each word may introduce noise in the process of the gloss-similarity score model as they may select not-optimal glosses.


\subsection{Candidate Ranking Task}

We also evaluate LexSubCon in the candidate ranking task for both the LS07 an CoInCo dataset.
In this sub-task the candidate substitution words are provided and the main task of the system is to create the most appropriate ranking of the candidates. 
 \begin{table}[h!]
 \centering
  \begin{tabular}{l  c c  }
  \hline

\textbf{Method}   & LS07  &    CoInCo     \\ \hline
LexSubCon    & \textbf{60.6}  & \textbf{58.0}    \\
\textit{-w/o Pr.} & 58.8& 56.3  \\
\textit{-w/o Gl.} & 60.3& 57.4  \\
\textit{-w/o Sen. } & 59.8 & 57.1 \\
\textit{-w/o Val.}& 56.8& 53.8\\
  Bert$_{s_p, s_u}$* & 58.6 &  55.2 \\  
LexSubCon (trial+test)  & 60.3  & 58.0   \\
  Bert$_{s_p, s_u}$* (trial+test) & 57.9 &  55.5 \\

 XLNet+embs &57.3 & 54.8 \\

 context2vec  &56.0&47.9\\
  Trans. learning &51.9 &- \\
   Sup. learning  &55.0 &- \\
    PIC  &52.4&48.3\\
    Substitute vector &55.1& 50.2 \\
    Addcos &52.9&48.3 \\
Vect. space mod.  & 52.5 & 47.8 \\\hline

  \end{tabular}
  \caption{Comparison of GAP scores (\%) in previous published results in the candidate ranking task of our implementation  of LexSubCon and  Bert$_{s_p, s_u}$ \cite{zhou-etal-2019-bert}. We also provide the results on the entire dataset with (trial+test).  Models: XLNet+embs \cite{compa_study},   Context2vec \cite{melamud-etal-2016-context2vec}, Transfer learning \cite{hintz-biemann-2016-language},  Supervised learning\cite{szarvas-etal-2013-learning},
  PIC \cite{pic}, Substitute vector \cite{melamud-etal-2015-modeling},  Addcos \cite{melamud-etal-2015-simple}, Vector space modeling \cite{kremer-etal-2014-substitutes}. 
  }
  \label{tab:gap_table}
\end{table}

   \begin{table*}[h!]
\begin{tabular}{ |c|c|c|c|c|} 
\hline
Word &Sentence   & Gold Ranking  & LexSubCon & BERT$\textsubscript{based}$ \\
\hline
terrible & ..have a \textbf{terrible} effect &awful, very bad, appalling,  &horrible,  & negative,\\
& on the economy & negative, formidable   &horrific, awful & major, positive \\ \hline
return &  ..has been allowed to  &  go back, revert, &revert, retrovert, & recover, go, \\
&  \textbf{return} to its wild state  & resume, regress &regress&restore \\ \hline

\end{tabular}
\caption{Examples of target words and their top lexical substitutes proposed by LexSubCon and BERT$\textsubscript{based}$ model.}
\label{tab:example}
 \end{table*}

Table \ref{tab:gap_table} provides the evaluation results in the candidate ranking task of LexSubCon and of the previous state-of-the art models. We report the results both on the test set and on the entire dataset (trial+test), 
in order to have a complete comparison as some of the previous state of the art models were evaluated on the entire datasets  and some were evaluated only in the testing portion of the datasets.
It can be observed that all the features have a positive effect on the performance of LexSubCon thus allowing it to outperform the previous state-of-the-art methods. Specifically, the results  demonstrate the positive effect of the features on accurately ranking a list of potential candidates as LexSubCon outperforms the previous methods even in the scenario where they are all provided with the same substitution candidate list.

\subsection{ Qualitative Substitution Comparison}

In Table \ref{tab:example}, we provide different examples of   target words and their top lexical substitutes proposed by LexSubCon and the  BERT$\textsubscript{based}$ model in order to demonstrate the effect of external lexical resources  on the performance of a contextual model. As it can be observed,
for the target word \textit{terrible}, the BERT$\textsubscript{based}$ model proposes a candidate word (\textit{positive}) which may fit in the sentence but has the opposite meaning of the target word. However, LexSubCon  provides semantically similar candidates by using information from different signals (e.g comparison of the definition of each word). 
In addition, for the target word \textit{return}, our model  identifies appropriate candidates that are not in the vocabulary of the contextual model (the word \textit{regress})  by introducing candidates from a external lexical database. These examples showcase that enriching  contextual models with external lexical  knowledge  can assist the model to provide more accurate candidates.

\section{Extrinsic Evaluation: Data Augmentation}
\label{extrinsic}
We evaluate the performance of  LexSubCon in the context of textual data augmentation. Specifically,
we conduct experiments on a popular benchmark text classification tasks of the English subjectivity/objectivity dataset (SUBJ) \cite{10.3115/1218955.1218990}\footnote{license: https://tinyurl.com/t-license}. The SUBJ dataset contains 5000 subjective and 5000 objective processed sentences (movie reviews). We train the LSTM model (with the same hyperparameters)  which was used in \cite{Wei2019EDAED} to measure the effect of different data augmentation techniques.   We compare our method with previous state-of-the-art lexical substitution models and with other popular textual data augmentation techniques:  \begin{enumerate*}[label=(\roman*)] \item the back-translation technique (described in section \ref{sec:back})
\item the EDA framework \cite{Wei2019EDAED} which utilizes  four  operations of  Synonym Replacement and Random Insertion/Swap/Deletion in order to create new text.
\end{enumerate*}
Following the data generation algorithm in \cite{compa_study}, LexSubCon  creates new examples by sampling one word for each sentence,  generating the appropriate substitute list for this word and sampling one substitute with probabilities corresponding to their   substitute scores (which were normalized by dividing them by their sum)  to replace the original word with the sampled substitute.

\begin{figure}[h]
        \centering
        \includegraphics[width=\linewidth]{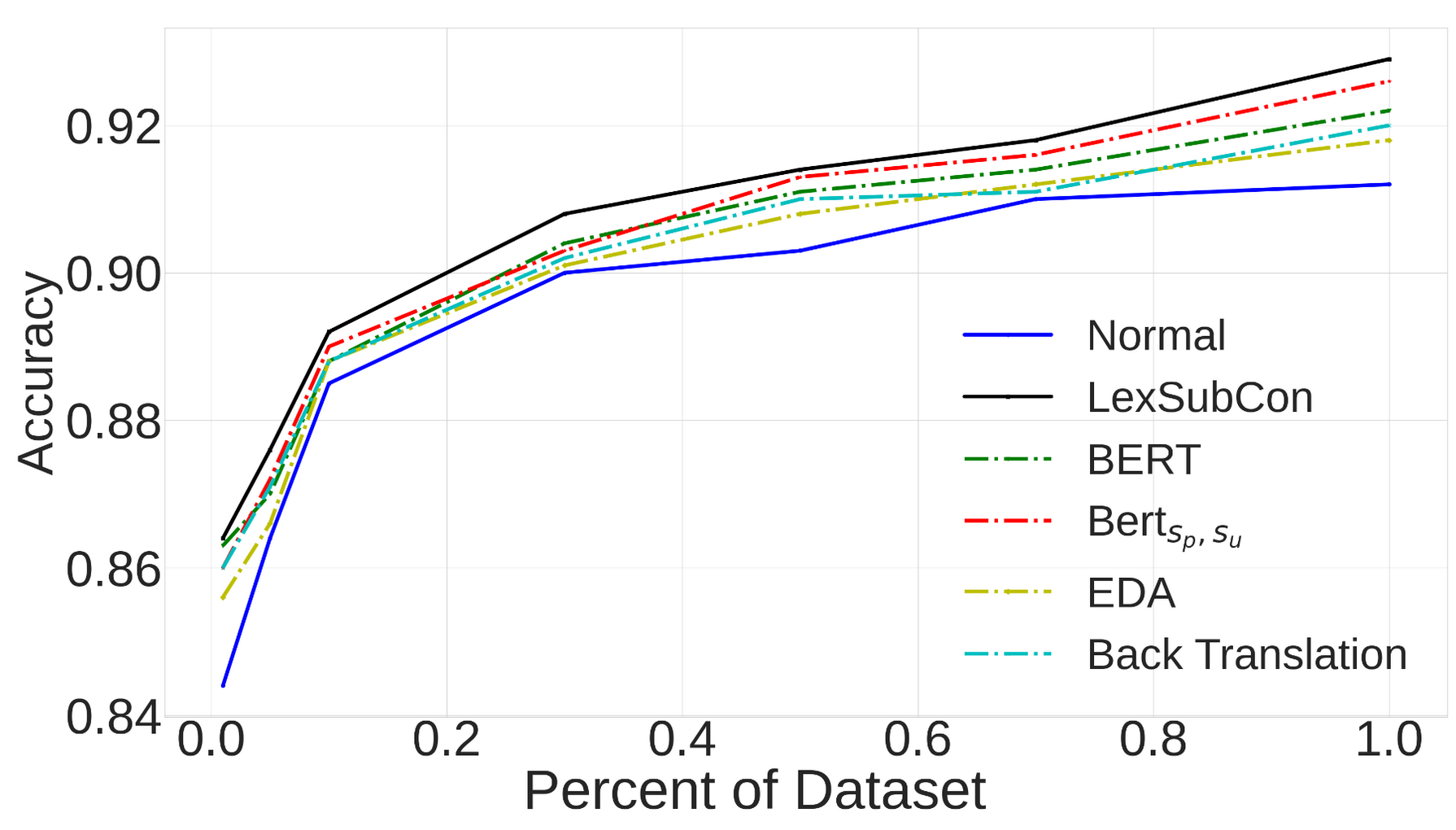}
    \caption{Accuracy with different train sizes  for different text augmentation techniques on the SUBJ dataset.}
    \label{fig:2}
\end{figure}
Figure \ref{fig:2} demonstrates how data augmentation   affects the classification depending on the size of the training set \cite{compa_study,Wei2019EDAED}. As it is expected the effect of each text augmentation technique on the performance of the model becomes more significant while the size of the train set is reduced.  Figure \ref{fig:2} also shows that the data   created with lexical substitution have a more positive effect to the performance of the model  than the other data augmentation techniques since back translation techniques may provide  text that does not follow  the syntactic rules of the target language and the EDA framework may create examples that could confuse the model by changing the  structure of the sentence due the random insertion and swapping of words. Finally, since LexSubCon can create more accurate substitution candidates than the standard BERT model and the  Bert$_{s_p, s_u}$ model, the  texts that were created with LexSubCon  have a more positive effect on the model's performance.

\section{Conclusion}
\label{conlusion}
This paper presents LexSubCon, an end-to-end lexical substitution framework based on contextual embedding models. LexSubCon establishes a new mix-up embedding strategy that outperforms the previous SOTA strategy of word embedding dropout for the   embedding of the target word  for the task of predicting accurate candidate words. LexSubCon  introduces the combined usage of  features from contextual embedding models and external lexical knowledge bases in order to calculate accurately the semantic similarity between a target word and its candidates.  We confirm that these features can improve the LexSubCon's performance as it outperforms other state-of-the-art results on two   benchmark datasets.

As for future work, we plan to address the limitations of this study including: \begin{enumerate*}[label=(\roman*)] \item examining the effect of using other models as the basis of our features (e.g. Albert \cite{DBLP:conf/iclr/LanCGGSS20});
\item exploring other candidate features for the ranking of the candidates (e.g. parser information \cite{szarvas-etal-2013-supervised})
 \item testing LexSubCon in  datasets of other languages using multi-language lexical databases  (e.g. MultiWordNet \cite{MultiWordNet} or BalkaNet \cite{Oflazer2001BALKANETAM}) to investigate further the model's general availability.
\end{enumerate*}


\section*{Ethical Consideration}
Lexical substitution can be useful in various natural language  processing (NLP) tasks such as textual data augmentation, paraphrase generation and text simplification. The results that we present in this paper  suggest that contextual word embeddings models, such as our framework (LexSubCon), can be a valuable tool for providing accurate substitution candidates that can be further used in a variety of down-stream tasks. 


We believe that there are many benefits of using   our   contextual embeddings models. For example, LexSubCon can be used as a data augmentation tool to provide artificial training data for tasks where the lack of sufficient training data may hurt the performance of the model. However, there are potential risks of  over-relying on any lexical substitution tool. Particularly, a lexical substitution model  can unintentionally change the  meaning of the original text thus leading to erroneous conclusions.

\bibliography{anthology,custom}
\bibliographystyle{acl_natbib}

\appendix



\end{document}